\documentclass[letterpaper, 10 pt, journal, twoside]{IEEEtran}  
\IEEEoverridecommandlockouts                              

\usepackage{cite}
\usepackage{graphicx}
\usepackage{amssymb}
\usepackage{amsmath}
\usepackage{algorithmic}
\usepackage{array}
\usepackage{mathtools}
\usepackage{multirow}
\usepackage{makecell}
\usepackage{caption}
\usepackage{subcaption}
\usepackage{booktabs}
\usepackage{hyperref}
\usepackage{cleveref}
\usepackage{url}
\usepackage{color, colortbl}

\graphicspath{{./images/}}
\DeclareGraphicsExtensions{.pdf,.eps}
\newcommand{\egmail}[6]{\{\href{mailto:#1@alibaba-inc.com}{#1},\href{mailto:#2@alibaba-inc.com}{#2},\href{mailto:#3@alibaba-inc.com}{#3},\href{mailto:#4@alibaba-inc.com}{#4},\href{mailto:#5@alibaba-inc.com}{#5},\href{mailto:#6@alibaba-inc.com}{#6}\} @alibaba-inc.com}

\newcommand{\ctrlpt}[1]{\mathbf{C}_{#1}}                
\newcommand{\proj}[1]{\pi({#1})}                        
\newcommand{\tension}{\tau}                             
\newcommand{\mahadist}[2]{\left\|{#1}\right\|_{#2}^2}   

\title{\LARGE \bf
Road Mapping and Localization using Sparse Semantic Visual Features
}
\author{Wentao~Cheng,
        Sheng~Yang,
        Maomin~Zhou,
        Ziyuan~Liu,
        Yiming~Chen,
        Mingyang~Li%
\thanks{Manuscript received: October 15, 2020; Revised: January 15, 2021; Accepted: February 16, 2021.}
\thanks{ This paper was recommended for publication by
	Editor Eric Marchand upon evaluation of the Associate Editor and Reviewers' comments.}
\thanks{All authors are with the Alibaba Group, Hangzhou, China. E-mail:
\egmail{wentaocheng}{shengyang}{maomin.zmm}{yuan.lzy}{yimingchen}{mingyangli}. W. Cheng and S. Yang contributed equally.}
\thanks{Digital Object Identifier (DOI): see top of this page.}}

\begin{document}
\maketitle

\markboth{IEEE Robotics and Automation Letters. Preprint Version. Accepted March, 2021}
{CHENG \MakeLowercase{\textit{et al.}}: Road Mapping and Localization using Sparse Semantic Visual Features}

\begin{abstract}
We present a novel method for visual mapping and localization for autonomous vehicles, by extracting, modeling, and optimizing semantic road elements. Specifically, our method integrates cascaded deep models to detect standardized road elements instead of traditional point features, to seek for improved pose accuracy and map representation compactness.
To utilize the structural features, we model road lights and signs by their representative deep keypoints for skeleton and boundary, and parameterize lanes via piecewise cubic splines. 
Based on the road semantic features, we build a complete pipeline for
mapping and localization, which includes a) image processing front-end, b) sensor fusion strategies, and c) optimization back-end.
Experiments on public datasets and our testing platform have demonstrated the effectiveness and advantages of our method by outperforming traditional approaches.
\end{abstract}

\section{Introduction}
\label{sec:intro}


To enable autonomous vehicles to fully navigate across urban environments, one of the key technical component is mapping and localization, which provides high-fidelity pose estimates~\cite{cadena2016past}. 
In general, the design guidelines for mapping and localization can be summarized as: accurate, robust, and efficient. The first two properties are to ensure the safety of autonomous vehicles under varying environments, and the last one is to allow the low-cost usage of processor, memory, and storage to make autonomous vehicles more affordable.

In this work, we focus on mapping and localization primarily using visual sensors, which are typically of low prices, low power consumptions, and widely utilized in current commercial vehicles. To date, most visual maps consist of 3D distinctive points, computed by probabilistically aggregating and optimizing information from point features detected from input images~\cite{mur2017visual,campos2020orb}. However, although point feature based algorithms are widely explored in recent years, they still suffer from the weakness in efficiency and robustness~\cite{rosinol2020kimera}. On one hand, those algorithms typically extract hundreds of point features per image, and compute millions or thousands of millions of points for town-size maps. This requires prohibitive amount of computational resources for both offline computing and online query process~\cite{schneider2018maplab}. On the other hand, traditional point features might not always be suitable for long-term persistent localization, since they might be difficult to be consistently recognized across different light/weather conditions or might be of high probability of disappearance~\cite{sattler2018benchmarking}.

To tackle the above problems, one solution is to constrain map elements to be both `persistent' and `compact' when building a map. Specifically, map elements should be persistent for {\em valid} registration, where dynamic elements are filtered, and stable features are retained. Additionally, map elements need to be sparse and frequently observable to achieve compactness, whereas stored targets are highly query-able for {\em efficient} registration. To achieve those properties, instead of primarily relying on geometrical features as traditional methods do~\cite{mur2017visual,campos2020orb,zhang2019localization}, one can consider both geometrical and semantical attributes in feature representation.

In existing mapping and localization methods, 
high-level feature representations have been designed to enhance the estimation performance, by enriching applicable geometric components of metric maps. Representative methods include the ones that utilize lines~\cite{gomez2019pl}, planes~\cite{yang2016pop}, and boxes~\cite{yang2019cubeslam}. Inspired by recent success in computer vision algorithms for semantic understanding, detecting semantic components in mapping and localization is also under active investigation. 
\cite{kendall2015posenet} proposed a method to encode semantic information via a neural network to conduct end-to-end 6-DoF positioning. However, this method lacks of pose uncertainty guarantees and thus cannot be used directly for applications that require high safety. Alternatively, 
~\cite{schonberger2018semantic} introduced an algorithm for semantic descriptor learning, which however does not meet the demands of low-cost computing.

In autonomous driving applications, we notice that common roads naturally consist of persistent and compact elements: standardized traffic signs (hung or painted on the ground), lanes, light poles, and so on. Therefore, to facilitate stable and low-cost autonomous driving systems, we propose to detect and parameterize road elements and design a novel semantic mapping and localization pipeline. We view this pipeline the key contribution of this work, which includes:
\begin{itemize}
\item A convolutional neural network (CNN) backed image processing front-end to extract semantic features.
\item Methods on parameterizing road elements and designing loss functions. 
\item Semantic optimization modules that can be used for both offline mapping and online localization.
\end{itemize}

We notice that there exist methods that are conceptually similar to our method, by segmenting road images and selecting points in stable regions, e.g.,~\cite{kaneko2018mask}. However, point features in stable semantic regions might not necessarily be stable and compact, and high level information, e.g., curves, is not utilized. By contrast, our method utilizes multi-source semantic information and provides more compact representation, reaching better `persistence' and `compactness'.
\section{Semantic Mapping and Localization}
\label{sec:method}

\subsection{System Overview}
\label{sec:method:overview}

\begin{figure}[ht]
  \centering
  \includegraphics[width=\linewidth]{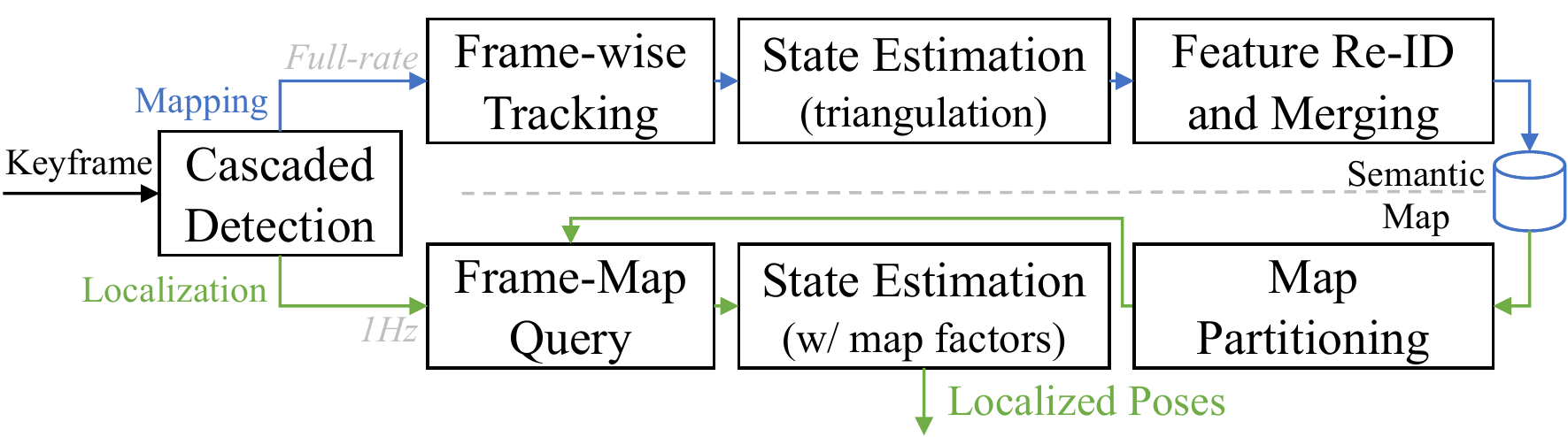}
  \caption{Overview of the proposed semantic mapping and localization pipeline. The blue arrow connects blocks in mapping stage to build compact maps offline, and the green arrows indicate the data flow of online localization stage.}
  \label{fig:pipeline}
\end{figure}

The backbone of our mapping and localization system is a tightly-coupled state optimization framework with both batch and sliding window strategies~\cite{mur2017orb,qin2018vins,schneider2018maplab}, as shown in Fig.~\ref{fig:pipeline}. 
Specifically, 
our algorithm proposes to build semantic maps based on standardized road entities offline, and use such maps for online localization. The involved semantic entities include three major types: horizontal objects, ground objects, and lanes (Sec.~\ref{sec:method:choose}). Giving a keyframe, the perception module performs cascaded deep detection to extract instances and their representative points as visual features (Sec.~\ref{sec:method:detect}). 

During offline mapping, the perception module is executed for every keyframe. Detected results between sequential keyframes then are tracked (Sec.~\ref{sec:method:tracking}), to establish multi-view associations for jointly estimating camera trajectory and landmark poses (Sec.~\ref{sec:method:estimator}). Subsequently, re-observed entities in previously visited road sections are re-identified and merged through loop detection (Sec.~\ref{sec:method:loop}). Finally, these optimized states are serialized as map assets for localization (sec.~\ref{sec:method:save}).
During online map-aided localization, the perception module runs at lower frequencies to allow low-cost consumptions on computational units.
Therefore, semantic features are obtained via a mixed detection and tracking strategy (Sec.~\ref{sec:method:reloc}). Those features are matched against the saved maps, and used by a sliding-window optimization based odometry system to reduce global drifts.

\subsection{Selection of Road Features}
\label{sec:method:choose}

\begin{figure}[ht]
  \centering
  \includegraphics[width=\linewidth]{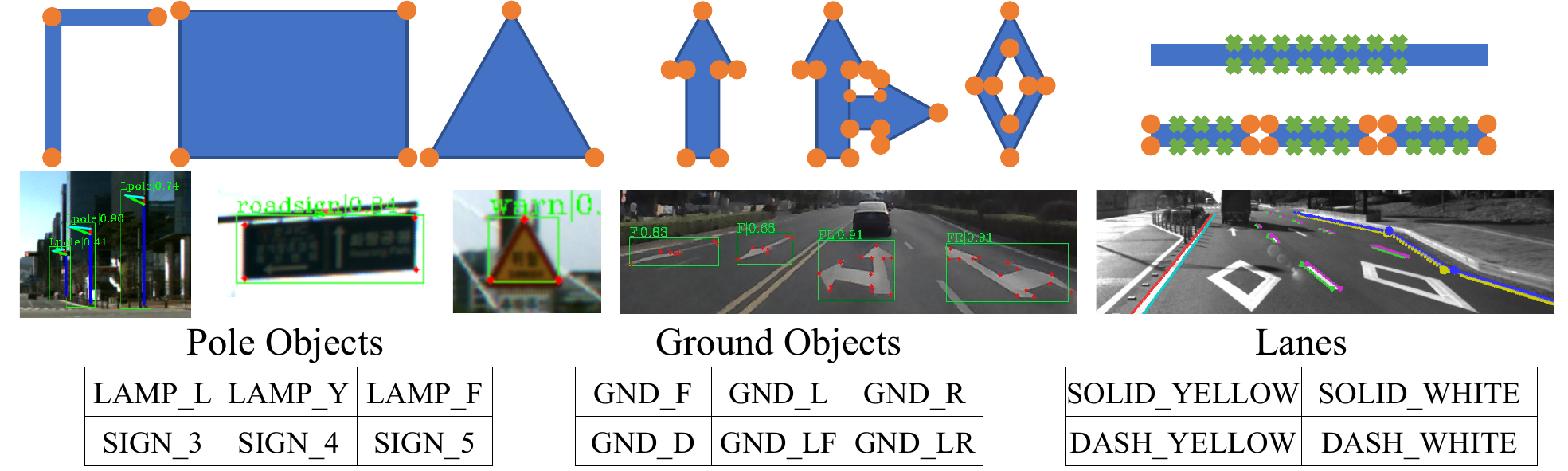}
  \caption{Semantic entities and their structures defined and used in our maps, with examples of detected instances. Orange: Indexed deep points. Green: Contour sample points.}
  \label{fig:selected_features}
\end{figure}

Considering map sparsification and query effectiveness, the following standardized targets on urban roads are suitable to be detected as semantic landmarks: 1) Lamps and traffic signs on top of the poles beside roads are stable and high enough to be captured by front-facing cameras.
2) Although sometimes occluded by vehicles, the ground area occupies nearly half of each image, making those high contrast signs painted on the ground unneglectable. 3) Similar to ground signs, painted solid and dashed lanes are also frequently observed. 
Solid lanes provide one directional motion constraints
and corners from dashed lanes can be considered as indexed point landmarks. In this work, 
we choose the above mentioned semantic types as target objects, as visualized in Fig.~\ref{fig:selected_features}, to build our semantic maps.

\subsection{Detection of Road Features}
\label{sec:method:detect}

Our two-stage cascaded detection module first performs instance level detection to get instances as {\em indexed representative pixels} on box (i.e., both pole and ground) objects and {\em sample pixels} on lane contours. Then, along these detected dash lanes, we evaluate $64 \times 64$ image patches to cascadingly detect {\em indexed dashed lane corners}. To reduce duplicated computations on shareable processes such as feature extraction, we refer to an anchor free detection method CenterNet~\cite{zhou2019objects}, which separates the low-level feature extraction process from the top-level heads, to make these heads adaptable to different tasks.

\textbf{Network design.} 
We use the DLA-34~\cite{yu2018deep} and the DCN module as the backbone for feature extraction. After deconvolution, we get a downsampled feature map for multiple heads $H$ with their loss functions listed in Table~\ref{tab:dl_loss}.

\begin{enumerate}
	\item \textbf{Object boxes and keypoints.} Following the human pose estimation task of the CenterNet~\cite{zhou2019objects}, our model supports a mixed detection of multiple object boxes and keypoints with six proposed heads.
	\item \textbf{Lanes.} 
	The boundary regression head $H_{7}$ outputs the left-right and top-bottom boundary points relative to each foreground point. The output type head $H_{8}$ is a three-category classification branch, which determines the left-right or top-bottom (or neither) boundaries of a lane point. We also integrate embedding head $H_{9}$ proposed by LaneNet~\cite{neven2018towards} for reducing the dimension of feature representation of each location. These feature representations are later clustered via DBSCAN, and thereby different lane types can be distinguished. 
	\item \textbf{Dashed lane corners.}	Given $64 \times 64$ image patches, we detect corners in pairs for generating a line segment. To compare with annotations,
	we add two heads $H_{10}$ and $H_{11}$ to compute angular and length differences, respectively.
\end{enumerate}

\begin{table}[ht]
  \centering
  \caption{Loss definition and weight $w$ of different heads $H$ proposed for tasks 1) - 3) listed above.}
  \setlength{\tabcolsep}{2pt}
  \begin{tabular}{l|l|c|cccc}
    \hline
    Note          & Loss Definition                                                  & $w$    & 1)      & 2)      & 3)      \\ \hline
    $H_{1}$      & Center: L1 Focal loss                                             & 1.0    & $\surd$ & $\surd$ & $\surd$ \\
    $H_{2}$      & Center: L1 offset by the output stride                            & 1.0    & $\surd$ &         & $\surd$ \\
    $H_{3}$      & Box: L1 loss of box width and height                              & 0.1    & $\surd$ &         &         \\
    $H_{4}$      & Indexed point: L1 Focal loss                                      & 1.0    & $\surd$ &         & $\surd$ \\
    $H_{5}$      & Indexed point: L1 offset by the output stride                     & 1.0    & $\surd$ &         & $\surd$ \\
    $H_{6}$      & Indexed point: L1 offset to the center point                      & 1.0    & $\surd$ &         &         \\
    $H_{7}$      & Lane: L1 distance to four sub-boundaries                          & 0.2    &         & $\surd$ &         \\
    $H_{8}$      & Lane: Cross entropy loss for sub-boundary types                   & 1.0    &         & $\surd$ &         \\
    $H_{9}$      & Lane: Instance embedding loss~\cite{neven2018towards}             & 1.0    &         & $\surd$ &         \\
    $H_{10}$     & Line Segment: L1 angular difference                               & 1.0    &         &         & $\surd$ \\
    $H_{11}$     & Line Segment: L1 length difference                                & 1.0    &         &         & $\surd$ \\
    \hline
  \end{tabular}
  \label{tab:dl_loss}
\end{table}

\subsection{Feature Tracking for Semantic Entities}
\label{sec:method:tracking}

Given two consecutively detected frames, the tracking module first accumulates their relative transformation $\mathbf{T}'$ via integrating IMU measurements~\cite{zhang2019vision}. For ground objects including signs and lanes, we use the following reprojection equation to map a pixel $p$ onto the other frame as $p'$ as:
\begin{equation}
\begin{aligned}
p'            &= \pi\left(\mathbf{T}' \cdot \pi'(p, z_p) \right),\,
\text{with}\,\,\, \mathbf{G} \cdot \pi'(p, z_p) = 0
\label{equ:reproj}
\end{aligned}
\end{equation}
where $z_p$ is the depth of pixel $p$ considered valid if positive. $\pi(\cdot)$ and $\pi'(\cdot,z)$ are the camera projection and back-projection operators respectively. We use $\mathbf G = 
[\theta, \phi, d]$ to denote a plane in the camera frame, and $\mathbf{G} \cdot \mathbf{x} \triangleq [ \cos{\theta} \cos{\phi}, \cos{\theta} \sin{\phi}, \sin{\theta} ]^\top \mathbf{x} + d = 0$. Such coefficients are solved jointly the state estimation as shown in Sec.~\ref{sec:method:estimator}.

We use the Hungarian matching strategy to associate ground features in both {\em instance-wise} and {\em pixel-wise} manners in the pixel space: 1) During {\em instance-wise} association, we compute the Intersection-over-Union (IoU), on both polygons for regular objects and 5.0 pixel width polylines for lanes. 2) During {\em pixel-wise} association, we compute the reprojected pixel distance for their indexed keypoints. Matches with IoU percentage $<50\%$ and pixel distance $> 5.0$ are ignored.

For keypoints detected in vertical objects such as poles, we use the optical flow method~\cite{lucas1981iterative} for tracking between frames.
During feature tracking, we keep classical keypoints extracted, described, and tracked by the GFTT extractor~\cite{shi1994good} and the FREAK descriptor~\cite{alahi2012freak}, since they are not only a part of the visual-inertial odometry, but also stably tracked point features worthy of being included in structured objects. Unlike segmentation that outputs masks, detected 2D boxes might contain GFTT feature keypoints from the background areas, especially in the pole instances. Hence, we perform outlier rejection for these background feature keypoints during the state initialization discussed in Sec.~\ref{sec:method:estimator}.

\subsection{Representation and Initialization of Road Lanes}
\label{sec:method:curve}

We use the piecewise cubic Catmull-Rom spline curves~\cite{catmull1974class} to represent the left and right contour of each 3D lane through a series of control points $\mathbf{C}_k \in \mathbb{R}_3$. Every four continuous control points will determine the shape $\mathbf{C}(t') \in \mathbf{R}_3$, $t' \in [0, 1]$ between two middle points $\ctrlpt{k}$ and $\ctrlpt{k+1}$, as:
\begin{equation}
\begin{aligned}
\mathbf{C}(t') &=
\begin{bmatrix*}[l]
{t'}^3 \\
{t'}^2 \\
{t'}   \\
1
\end{bmatrix*}^\top
\begin{bmatrix*}[r]
-\tension & 2-\tension &  \tension-2 &  \tension \\
2\tension & \tension-3 & 3-2\tension & -\tension \\
-\tension &          0 &    \tension &         0 \\
        0 &          1 &           0 &         0 
\end{bmatrix*}
\begin{bmatrix*}[l]
\ctrlpt{k-1} \\
\ctrlpt{k}   \\
\ctrlpt{k+1} \\
\ctrlpt{k+2}
\end{bmatrix*}^\top,
\label{equ:curve}
\end{aligned}
\end{equation}
where the tension $\tension = 0.5$ describes how sharply such a curve bends at its control points. For notation simplicity of a piecewise spline curve, we denote $t \triangleq k + t' \in [2, M+1]$ for a whole piecewise curve $\mathbf{C}(t)$ consisting of $M$ segments controlled by $M+2$ points $\mathbf{C}_k, k \in [1, M+2]$. The first and last control points $\ctrlpt{1}$ and $\ctrlpt{M+2}$ are off-the-curve for adjusting the direction at its endpoints $\ctrlpt{2}$ and $\ctrlpt{M+1}$.

For initializing these control points, we first perform ray-ground intersection as Eq.~\ref{equ:reproj} of its detected sample and corner pixels, to get the point set $\bigcup_{c,j} \{\mathbf{P}_{cj}\}$ from multiple camera observations $c$. Then, we randomly pick $N$ samples among such a set as $\mathbf{C}_k, k \in [2, N+1]$ to examine the  losses:
\begin{equation}
\begin{aligned}
\mathop{argmin}\limits_{\mathbf{C}_k} & \sum_{c,j} \mahadist{\mathbf{P}_{cj} \!-\! \mathbf{C}(t_{cj})}{^{x}\Omega} \!+\! \sum_{k=3}^{N+1} \mahadist{ | \mathbf{C}_k \!-\! \mathbf{C}_{k-1}| \!-\! \lambda_1}{^{y}\Omega}, \\
with \ \ & \left\langle \mathbf{C}_1, \beta_1 \right\rangle \leftarrow \left\{
\begin{aligned}
|\mathbf{C}_2 - \mathbf{C}_1 | = |\mathbf{C}_3 - \mathbf{C}_2 | \\
\beta_1 \cdot \mathbf{N}_{2} = \tau(\mathbf{C}_3 - \mathbf{C}_1) \\
\end{aligned} \right.,
\label{equ:curve_init}
\end{aligned}
\end{equation}
where the first Mahalanobis term is the 3D point-to-curve residual considering all samples. To calculate the distance, we explicitly solve the parameter $t_{cj}$ through the equation $\mathbf{C}'(t_{cj})(\mathbf{C}(t_{cj}) \!-\! \mathbf{P}_{cj}) = 0$ piece by piece. The real roots of the quintic equation stands for possible stationary points, which are then compared with all integers between $[2, N+1]$ to pick up the global minimum point-to-curve residual.
The second regularization term ensures these control points to be sampled evenly. We use the parameter $\lambda_1 = 40 m$ for controlling density, and $^{x}\Omega = (0.1 m)^2 \mathbf{I}_3$, $^{y}\Omega = (0.25 \cdot \lambda_1)^2 \mathbf{I}_1$ for balancing their weights.

The off-the-curve point $\mathbf{C}_1$ (and analogously for $\mathbf{C}_{N+2}$) is initialized by solving the equation of $\left \langle \mathbf{C}_1, \beta_1 \right\rangle$ above. $\mathbf{N}_2$ is the 3D normal of $\mathbf{C}_2$ computed by searching its pixel neighborhoods intersected to the ground. Such equations mean to solve possible intersections of a sphere and a line. If we found more than one solutions, we accept the candidate $\mathbf{C}_1$ located at the opposite site of $\mathbf{C}_3$.

For repeatedly picking $N$ sequential samples as middle control points during such a sample-consensus initialization, we first randomly pick one sample, and recursively expand on its two sides by randomly picking another sample whose distance is between $[0.5\lambda_1, 1.5\lambda_1]$. We generate at most 500 random control point sequences for each spline, and evaluate the loss in Eq.~\ref{equ:curve_init} to pick up the best sequence. Compared to existing works using B-splines~\cite{Pedraza2009Extending} or polyb\'eziers~\cite{rao2012curveslam,wang2018fully}, Catmull-Rom splines cross through all middle control points for us to use such a random picking strategy for initialization.

\subsection{State Estimator Design}
\label{sec:method:estimator}

For clarity, we summarize the indexed notes used in our method in Table. \ref{tab:notes}. To add such semantic entities into a tightly-coupled visual-inertial odometry~\cite{zhang2019vision}, we introduce five new types of optimizable variables in our system summarized as following:

\begin{table}[ht]
	\centering
	\caption{Multiple indexing notes used in state estimation.}
	\begin{tabular}{l|l}
		\hline
		\multicolumn{2}{c}{$a$: ground and pole objects \quad $b$: splines \quad $c$: frames} \quad $j \in \{m, n\}$ \\ \hline
		$i$: pixels of $a$ as $(\cdot)^{c}_{a^i}$  & $k$: control points of $b$ as $(\cdot)_{b^k}$ \\ \hline
		$m$: samples of $b$ as $(\cdot)^{c}_{b^m}$ & $n$: corners of $b$ as $(\cdot)^{c}_{b^n}$    \\
		\hline
	\end{tabular}
	\label{tab:notes}
\end{table}

\begin{enumerate}
    \item The 3D location $\mathbf{P}_{a^i} \in \mathbb{R}_3$ of deep or classical keypoint $i$ within a ground or pole object $a$. We use the inverse depth parameterization~\cite{civera2008inverse} to optimize these points.
    \item The ground coefficients $\mathbf{G}$ in the camera coordinate as previously used in Eq.~\ref{equ:reproj}. We approximate observable regions of the ground within each frame as a plane. We use spherical local parameterization\footnote{We refer readers to the Plane3D implementation from g2o~\cite{grisetti2011g2o}.} to apply updates on its two angles. $\theta \in [0, \pi], \phi \in [0, 2\pi], d \in \mathbb{R}_3$.
    \item The vertical plane for a pole object $a$ in the global coordinate denoted as $\mathbf{V}_{a}(\varphi, e) \cdot \mathbf{x} \triangleq [\cos{\varphi}, \sin{\varphi}, 0]^\top \mathbf{x} + e = 0$, whose Z-axis is horizontal with the gravity direction after the visual-inertial odometry is initialized. $\varphi \in [0, 2\pi], e \in \mathbb{R}_3$.
    \item The control points $\mathbf{C}_{b^k} \in \mathbb{R}_3$ of a spline $b$, depicting the left or right contour of a road lane.
    \item The dynamic association parameters $t^{c}_{b^j} \in \mathbb{R}$ for associating pixel $p^{c}_{b^j}$ detected on a frame $c$ for a spline $b$. For clarity of notation, we use $j \in \{m, n\}$ for indexing sample and corner pixels, respectively.
\end{enumerate}

\begin{figure}[ht]
	\centering
	\includegraphics[width=\linewidth]{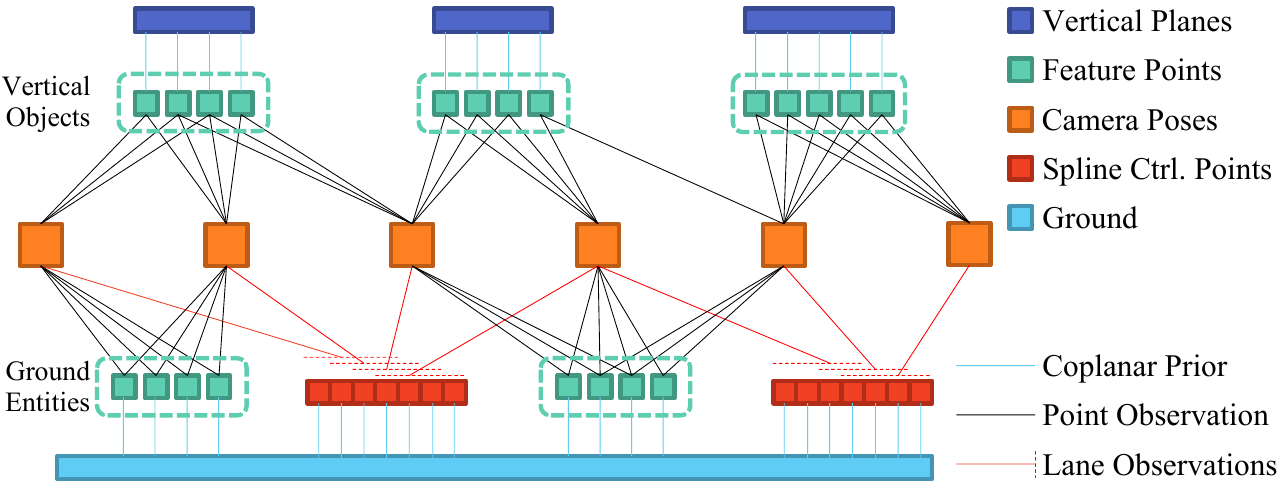}
	\caption{Illustration of the proposed factor graph structure. Basic graph elements from visual-inertial odometry~\cite{zhang2019vision} systems are hidden.}
	\label{fig:factor_graph}
\end{figure}

Based on the new variables above and the original frame poses $\mathbf{T}_c$ for each image frame $c$, we add three types of constraints (Fig.~\ref{fig:factor_graph}) based on the detected and tracked semantic features, including:

\begin{enumerate}
    \item \textbf{Points observation factors}. We tend to triangulate and parameterize regular keypoints as previous methods~\cite{mur2017orb}, through the following constraint:
\begin{equation}
^{1}\mathbf{f}^{c}_{a^i} = \mahadist{p^{c}_{a^i} - \proj{\mathbf{T}_{c}^{-1} \cdot {\mathbf{P}_{a^i}}}}{^{1}\Omega^{c}_{a^i}},
\label{equ:constraint_point}
\end{equation}
where $p^{c}_{a^i} \in \mathbb{R}_2$ is the location of the detected pixel. $^{1}\Omega^{c}_{a^i} = \sigma^2 \mathbf{I}_2$ is the noise covariance assigned w.r.t. the precision of detection $\sigma$. We use the reported median pixel error in Table~\ref{tab:perception} for assigning $\sigma$ for different types of semantic keypoints, and $1.0$ for GFTT keypoint considering its accurate detection.
    \item \textbf{Spline observation factors}. We use the constraint below to dynamically associate sample and corner pixels $p^{c}_{b^j}$ as the measurements of control points $\mathbf{C}_{b^k}$ from a spline $b$:
\begin{equation}
^{2}\mathbf{f}^{c}_{b^j} = \mahadist{p^{c}_{b^j} - \proj{\mathbf{T}_{c}^{-1} \cdot \mathbf{C}_{b}(t^{c}_{b^j}})}{^{2}\Omega^{c}_{b^j}},
\label{equ:constraint_curve}
\end{equation}
where $t^{c}_{b^j}$ is the introduced jointly optimized dynamic association parameter, and $^{2}\Omega^{c}_{b^j}$ is also set according to the precision of detected samples and corners, respectively.
    \item \textbf{Coplanar prior factors}. Both vertical and horizontal coplanar priors are added to the optimization through the following form of residuals:
\begin{equation}
\begin{aligned}
       ^{3}\mathbf{f}_{a^i} &= \mahadist{\mathbf{V}_{a} \cdot \mathbf{P}_{a^i}}{^{3}\Omega_{a^i}}, \\
or \ \ ^{3}\mathbf{f}_{\mathbf{x}} &= \mahadist{\mathbf{G} \cdot (\mathbf{T}_{c}^{-1} \cdot \mathbf{P}^{c}_{\mathbf{x}}) }{^{3}\Omega^{c}_{\mathbf{x}}}, \mathbf{x} \in \{a^i, b^j\},
\end{aligned}
\label{equ:constraint_plane}
\end{equation}
where $^{3}\Omega = (0.4 m)^2 \mathbf{I}_1$ is assigned according to the thickness or noise of the corresponding planar objects. We assume that these observed lanes are locally planar within each camera view with consistent coefficients.
\end{enumerate}

\textbf{Initialization of ground and pole objects.} We initialize the above optimizable variables sequentially from 1) to 5):
For 1) when a GNSS-VIO trajectory is given, we triangulate these feature points from the estimated poses.
For 2), we use their contained deep points $\bigcup_i \{ \mathbf{P}_{a^i} \}$ to perform a 2D line fitting in the XOY plane.
After 2) has been initialized, we use a thickness criterion $| \mathbf{V}_{a} \cdot \mathbf{P}_{a^i} | < \sigma_1 = 0.3 m$ to reject background pixels contained in the detected 2D box. Similarly for 3), we transform each triangulated deep point of the ground to its observed image frame, and then perform 3D plane fitting for $\mathbf{G}$. If there are no ground signs detected for the ground initialization, we instead use the classical feature points within the convex-hull of detected lanes in each frame, and apply a RANSAC 3D plane fitting strategy~\cite{rusu20113d} to remove keypoints on the moving vehicles.

\textbf{Initialization of splines.}
Based on the solved ground coefficients $\mathbf{G}$, we use the method described in Sec.~\ref{sec:method:curve} to initialize 4). Finally for 5), we first assign $t^{c}_{b^j}$ with the solved 3D association parameter during spline initialization, and then perform a standalone non-linear optimization to refine. We filter those samples having multiple association hypotheses, i.e., exist sub-optimal association whose residual is less than 1.5 times of the best association. Compared to pixel-on-line whose projective association is deterministic as the perpendicular feet in the image space, explicitly formulating $t_{aci}$ from such a piecewise curve as the {\em argmin} of $t_{aci}$ in Eq.~\ref{equ:constraint_curve} is hard due to the piecewise representation and the projection operator $\pi(\cdot)$. So we add such a variable type during our state estimation to ensure the correctness of curve projective association. Specifically for those detected dashed lane corners because they are tracked pixel-wise, we associate multiple observations of a corner $p^{c}_{b^n}$ to one exact location $\mathbf{C}_{b}(t_{b^n})$ along the curve.

\textbf{Offline mapping case.} After these variables are initialized, we derive a factor graph optimization based on the common visual-inertial odometry constraints~\cite{zhang2019localization}, with Cauchy loss functions added on the above factors for numerical stability, and the pose of the first frame fixed as identity. All keyframes and detected entities are involved in the final bundle adjustment for solving poses and positions jointly.

\textbf{Online localization case.}
During online localization, we deserialize fixed semantic landmarks, i.e., control points of splines and regular 3D points, from a semantic map, and fix them in Eq.~\ref{equ:constraint_point} and~\ref{equ:constraint_curve}, to add constraints on relative poses between the camera and the map coordinate. Coplanar constraints as Eq.~\ref{equ:constraint_plane} are no longer required in this phase. We will further introduce how these factors are constructed through the proposed map query strategy in Sec.~\ref{sec:method:reloc}.

\subsection{Re-Identification and Feature Merging}
\label{sec:method:loop}

We perform 3D-3D association for re-identifying semantic objects rather than a frame-wise bag-of-words~\cite{galvez2012bags} query. The reason is that the density of duplicated objects (tens of meters) are relatively sparser than the localization uncertainty of GNSS-VIO odometry during mapping, and the visual appearances among these standardized road elements are too similar to be distinguished. During the instance-wise object and lane association, we regard triangulated objects with the distance between their centroids smaller than 5.0 meters (or 0.5 meters for lanes) as identical, and then cascadingly merge the observations of their contained deep and classical points in a Hungarian strategy: The semantic type of deep points is used for rejecting mismatches. While for each accompanied GFTT point, we use their FREAK descriptors among multiple frames for voting. We use the union-find algorithm to merge their observations and conduct another round of global state optimization.

\subsection{Data Structure of Semantic Maps}
\label{sec:method:save}

After an open-loop (Sec.~\ref{sec:method:estimator}) and a close-loop (Sec.~\ref{sec:method:loop}) bundle adjustment, we obtain the final frame poses and landmark positions. To save these assets for localization, we define the data structure of our semantic map $\mathcal{M}$ as the combination of observers $\mathcal{O}$ (i.e., frames used during mapping), landmarks $\mathcal{L}$, and a co-visibility graph $\mathcal{G}$ linking $\mathcal{O}$ and $\mathcal{L}$. For each observer $c$, we store its estimated pose in the global coordinate $\mathbf{T}_c$ and the closest GNSS measurement for online GPS queries. For each hierarchical semantic landmark, we store the semantic label and the 3D positions of contained deep and GFTT points. Neither FREAK descriptors nor frame-wise descriptors are stored in our semantic map.

\subsection{Localization Based on Semantic Maps}
\label{sec:method:reloc}

\begin{figure}[ht]
  \centering
  \includegraphics[width=\linewidth]{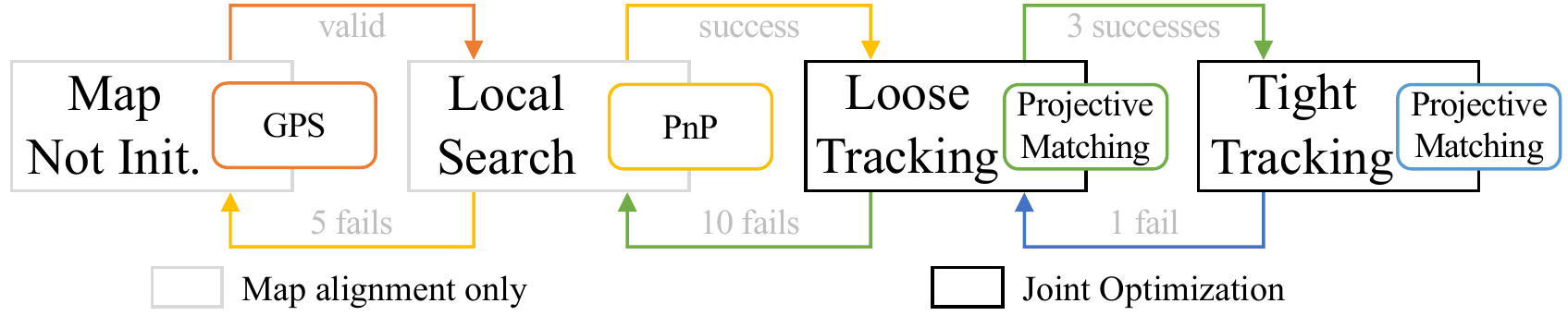}
  \caption{State machine during online localization.}
  \label{fig:state_machine}
\end{figure}

We use a state machine illustrated in Fig.~\ref{fig:state_machine} to assess the pose quality of online localization and perform different strategies accordingly. Starting from the {\em map uninitialized} state, in which the global transformation from the map coordinate to the current global coordinate $\mathbf{T}_{\mathcal{M}}$ is unknown, we use a coarse GPS measurement to retrieve the corresponding map partition, to get the corresponding observers in $\mathcal{O}_{\mathcal{M}} \subset \mathcal{O}$ whose relative translation is less than 30 meters, and switch to the {\em local search} mode.

In the {\em local search} mode, we can obtain the landmark subset $\mathcal{L}_{\mathcal{M}} = \mathcal{G}(\mathcal{O}_\mathcal{M})$, for establishing 2D-3D PnP-Ransac association~\cite{lepetit2009epnp} through indexed deep object keypoints. At least 12 inliers from 5 different entities are required for accepting such an estimated PnP pose as valid, to switch into the two {\em tracking} states and initialize $\mathbf{T}_{\mathcal{M}}$.

We use two stages of the {\em tracking} state as {\em loose} and {\em tight}, to apply different thresholds according to their pose quality. In both {\em tracking} states, reconstructed associations will be added to the sliding window optimization through Eq.~\ref{equ:constraint_point} and~\ref{equ:constraint_curve}, where the original $\mathbf{T}_c$ for the frame to global transformation is replaced by $\mathbf{T}_{c\mathcal{M}}$ for the frame to map transformation, and $\mathbf{T}_{\mathcal{M}}$ is regarded as an additional optimizable variable during the optimization.

In such two {\em tracking} modes, we project all map entities to the online detected frame $c$, and use the following criterion to accept a projective association through a Hungarian matching, as:
\begin{equation}
\begin{aligned}
^{\mathbf{y}}\mathbf{f}^{c}_{\mathbf{x}} < \mahadist{\sigma_{2}}{^{\mathbf{y}}\Omega^{c}_{\mathbf{x}}}, \ \mathbf{x} \in \{a^i, b^j\}, \ \mathbf{y} \in \{1, 2\},
\end{aligned}
\label{equ:proj_asso}
\end{equation}
where threshold $\sigma_{2}$ in the {\em tight} mode is set as 20.0 and 3.0 pixels for deep and classical keypoints, respectively, due to their density on the projected image. In the {\em loose} mode, we loose $\sigma_{2}$ to 30.0 pixels for labeled deep keypoints, and disable classical keypoints since they are not distinguishable when the geometry is not accurate. To remove the influence of association ambiguity, we reject the association to a map point if there exists another sub-optimal candidate whose projected distance $\hat{\mathbf{f}} / \mathbf{f} < 4.0$. For state transformation, we regard a projective matching as stable if more than 4 associations are constructed. 

\section{Experimental Evaluation}
\label{sec:eval}

\subsection{Datasets}
\label{sec:eval:dataset}

We use two real-world datasets to evaluate our performance, including a publicly available dataset KAIST~\cite{jeong2019complex}, and a self-recorded dataset. For KAIST sequences, we use the left camera with IMU and GNSS measurements, on urban road sequences 26, 28, 38, 39. Note that some of their trajectories are overlapping with others.
We also recorded two sequences in Hangzhou, from hardware synchronized sensors including a 10Hz ON AR0144 stereo camera, a 200Hz Bosch BMI 088 IMU and a 10Hz u-blox F9 RTK-GPS. We recorded a long sequence called Hangzhou-1 traveling 4.7 km for mapping, and another sequence Hangzhou-2 as a subset having 3.2 km for cross localization.

\subsection{Training and Performance of Perception Model}
\label{sec:eval:perception}

\begin{figure*}[ht]
	\centering
	\begin{subfigure}[b]{0.195\linewidth}
		\centering
		\includegraphics[width=\linewidth,trim=12 15 25 5,clip]{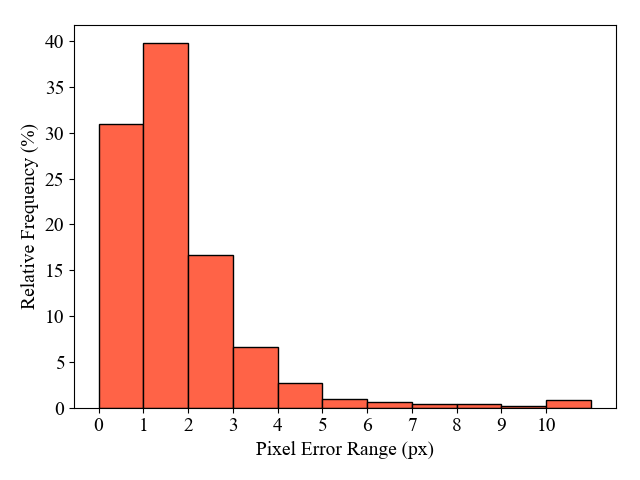}
		\caption{Poles}
	\end{subfigure}
	\begin{subfigure}[b]{0.195\linewidth}
		\centering
		\includegraphics[width=\linewidth,trim=12 15 25 5,clip]{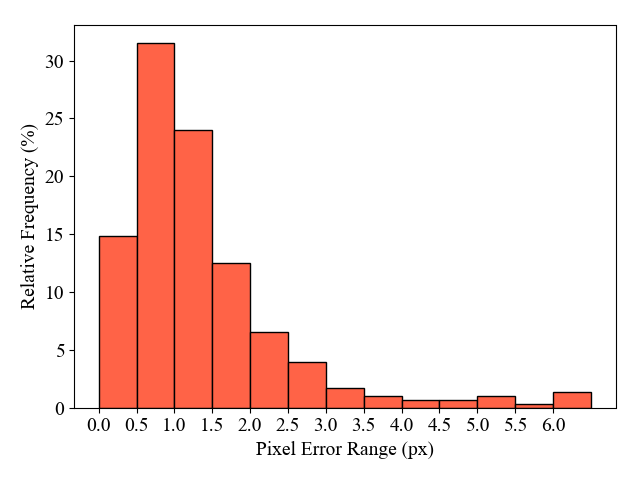}
		\caption{Traffic signs}
	\end{subfigure}
	\begin{subfigure}[b]{0.195\linewidth}
		\centering
		\includegraphics[width=\linewidth,trim=12 15 25 5,clip]{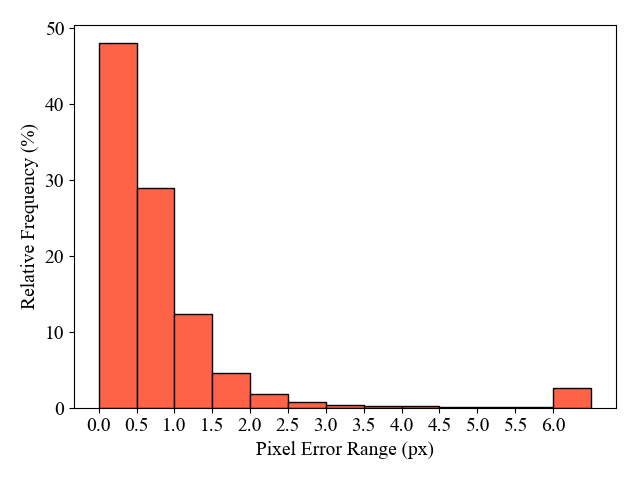}
		\caption{Ground signs}
	\end{subfigure}
	\begin{subfigure}[b]{0.195\linewidth}
		\centering
		\includegraphics[width=\linewidth,trim=12 15 25 5,clip]{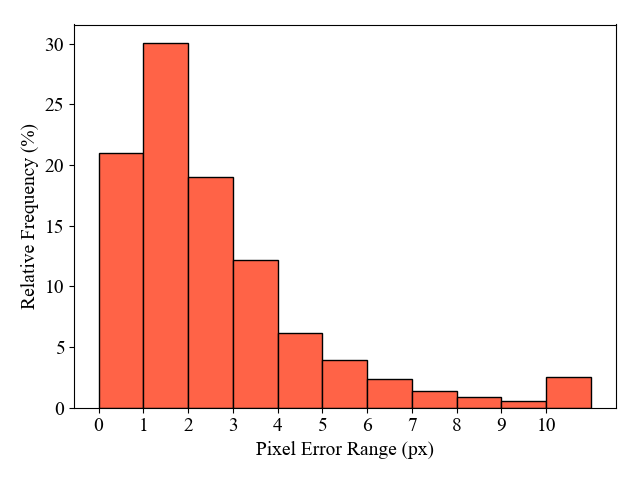}
		\caption{Lane samples}
	\end{subfigure}
	\begin{subfigure}[b]{0.195\linewidth}
		\centering
		\includegraphics[width=\linewidth,trim=12 15 25 5,clip]{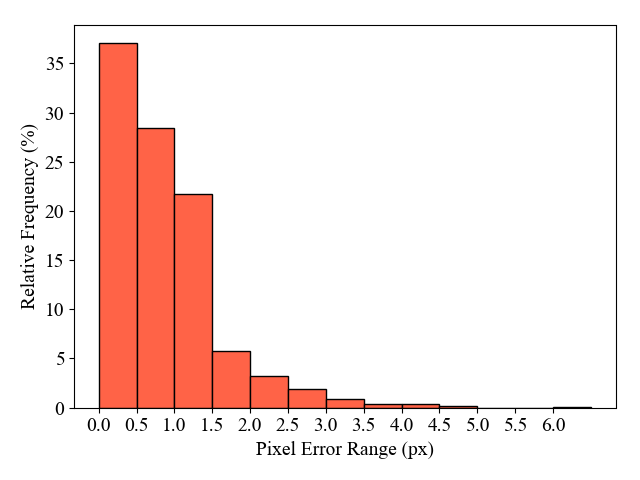}
		\caption{Lane corners}
	\end{subfigure}
	\caption{Pixel error histograms of detected keypoints on different categories, with exceeded errors trimmed into the last bin.}
	\label{fig:detection_accuracy}
\end{figure*}

\textbf{Training.} We manually annotated 3207 images extracted from these four KAIST sequences (4.4\% of all). These annotations include 2D object boxes, lane contours, and instance keypoints illustrated in Fig.~\ref{fig:selected_features}. We randomly divide labeled images into a training set and a test set with the ratio of 85\% and 15\%, and augment them through scaling and color enhancement to generate $512 \times 512$ model inputs. We use the Adam optimizer~\cite{kingma2014adam}, and set the initial learning rate to $1.25 \times 10^{-4}$, which drops by 0.1 at 100 and 200 epochs, to train the model separately in different tasks. For Hangzhou sequences, we similarly annotated 2467 images for detection.

\textbf{Performance.} Table~\ref{tab:perception} shows the performance of our trained deep model on the KAIST dataset, with box objects grouped by three representative sub-types (poles, traffic signs, and ground signs) due to their different appearances on images. We evaluate these tasks separately on the test set, to conclude their classification precision, detection recall, and additionally the pixel error of extracted pixels in Table~\ref{tab:perception}. We also report the distribution histogram of pixel error in Fig.~\ref{fig:detection_accuracy}.

\begin{table}[ht]
	\centering
	\caption{Performance of our proposed multi-task detection model. PRE.: Precision. REC.: Recall. APE.: Average pixel error. MPE.: Median pixel error. We use the MPE. for measurement uncertainties $\sigma$ in Eq.~\ref{equ:constraint_point} and~\ref{equ:constraint_curve}.}
	\begin{tabular}{l|ccc|cc}
	\hline
              & \multicolumn{3}{c|}{Keypoint in Box}    & \multicolumn{2}{c}{Lane}                  \\
		      &    Poles    & Signs       &      GND.   &  Samples        & Corners                 \\ \hline
	PRE.(\%)  & 85.9        & 89.0        & 93.3        &  93.3           & 98.1                    \\
	REC.(\%)  & 95.4        & 94.2        & 81.6        &  89.0           & 94.2                    \\
	APE.      & 3.809       & 2.237       & 1.689       &  2.973          & 0.857                   \\
	MPE.      & 1.952       & 1.405       & 1.077       &  0.524          & 0.717                   \\
	\hline
	\end{tabular}
	\label{tab:perception}
\end{table}

\begin{figure}[ht]
	\centering
	\includegraphics[width=\linewidth]{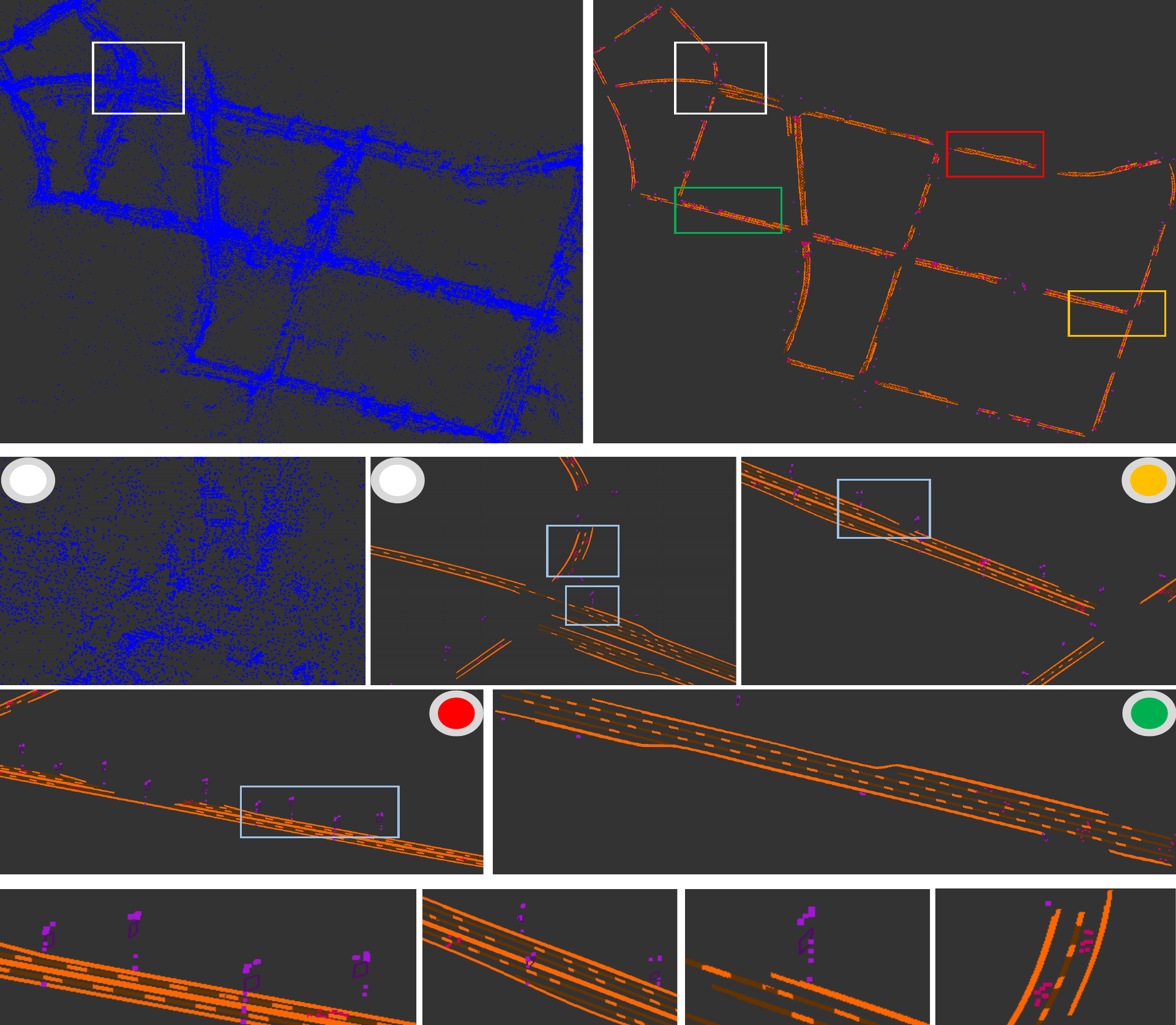}
	\caption{The semantic map of KAIST-38 with zoom-in views. Blue: Conventional map for reference. Orange: Road lanes. Purple: Points of road poles. Red: Points of ground signs.}
	\label{fig:teaser}
\end{figure}

\subsection{Evaluations on Localization and Mapping}
\label{sec:eval:map_loc}

\textbf{Experimental Settings.}
We choose to compare the proposed method with downsampled maps consisting of classical feature points described by FREAK~\cite{alahi2012freak}. About 500 GFTT~\cite{shi1994good} keypoints (same with our system) are extracted on each keyframe to perform sliding window optimization, frame-wise loop-closure, and final bundle adjustment.
We use high-quality GNSS measurements during offline mapping and low-quality GPS during online relocalization.
In detail, there are two differences between our proposed method and this baseline: 1) After the final bundle adjustment of offline mapping, we perform a spatial re-identification for semantic entities and thus require another round of final bundle adjustment. 2) During frame-to-map association in the online localization, our method replaces all classical frame-map local feature associations by these serialized semantic features stored in a semantic map.

To make the file size of these compared maps close, we use a set cover algorithm~\cite{lynen2015get} to downsample the classical feature map. It ensures that each member in the involved map observers $\mathcal{O}$ is able to observe at least a certain number of map points. Also, each FREAK descriptor is compressed into a 10-dimensional vector (40 bytes in total). By varying the parameter $K$ used in the set cover algorithm, we obtain several sparsified conventional maps with different memory footprints.
We also provide another variant of our method as a box-object-only version which does not contain any lane procedures for ablation study.
During the quantitative evaluation, we mainly test the absolute trajectory error by the translational root-mean-squared-error (T.RMSE)
~\cite{zhang2018tutorial}. For self and cross relocalization, we register online localization trajectory to the offline mapping trajectory for demonstrating the map-aided localization ability. Fig.~\ref{fig:teaser} visualizes a typical example of our semantic map.

\begin{figure}[ht]
	\centering
	\begin{subfigure}[b]{0.49\linewidth}
		\centering
		\includegraphics[width=\linewidth,trim=15 2 25 25,clip]{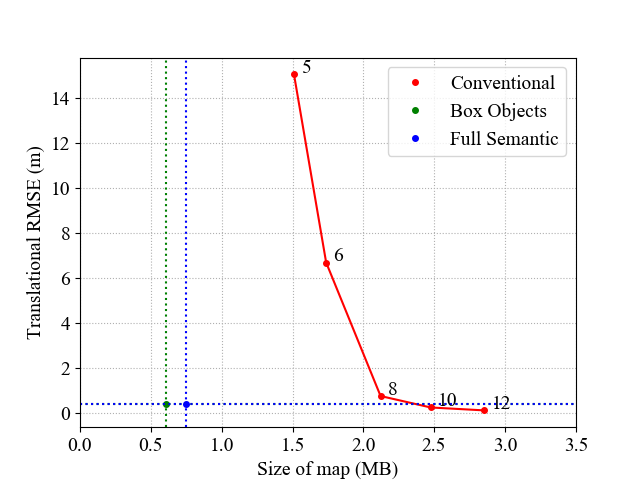}
		\caption{KAIST-26}
		\label{fig:26_result}
	\end{subfigure}
	\begin{subfigure}[b]{0.49\linewidth}
		\centering
		\includegraphics[width=\linewidth,trim=15 2 25 25,clip]{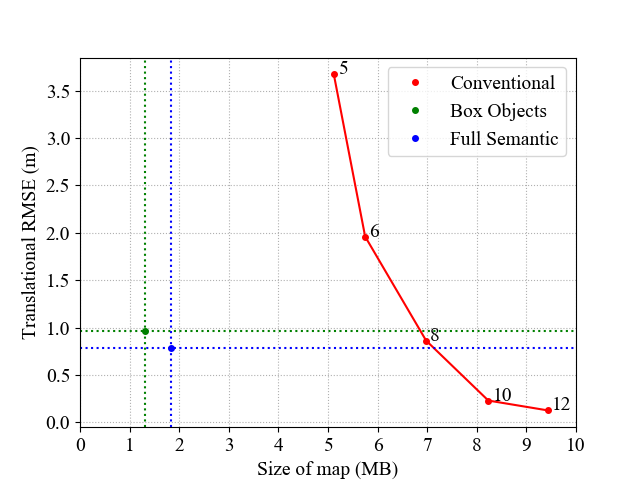}
		\caption{KAIST-28}
		\label{fig:28_result}
	\end{subfigure}
	\begin{subfigure}[b]{0.49\linewidth}
		\centering
		\includegraphics[width=\linewidth,trim=15 2 25 25,clip]{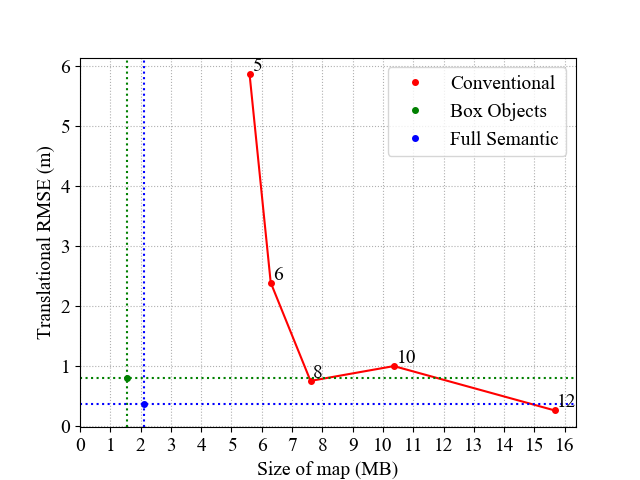}
		\caption{KAIST-38}
		\label{fig:38_result}
	\end{subfigure}
	\begin{subfigure}[b]{0.49\linewidth}
		\centering
		\includegraphics[width=\linewidth,trim=15 2 25 25,clip]{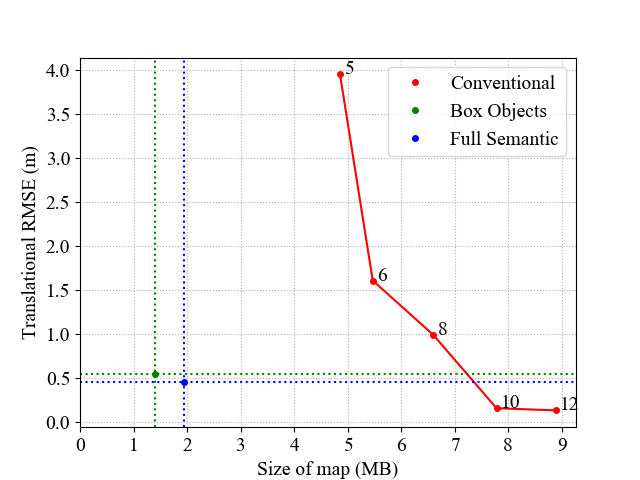}
		\caption{KAIST-39}
		\label{fig:39_result}
	\end{subfigure}
	\begin{subfigure}[b]{0.49\linewidth}
		\centering
		\includegraphics[width=\linewidth,trim=15 2 25 25,clip]{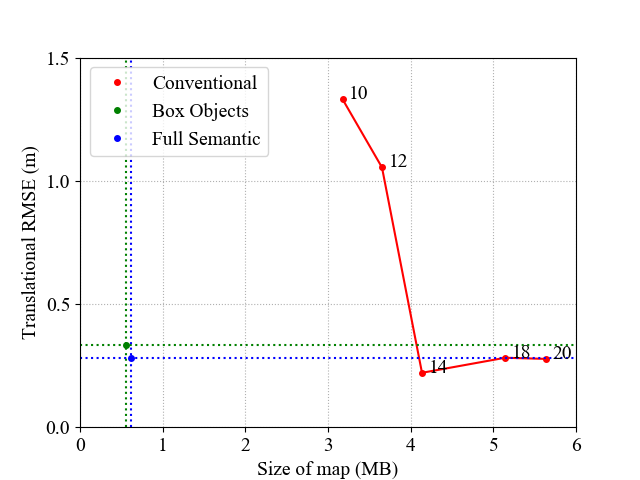}
		\caption{Hangzhou-1}
		\label{fig:09_result}
	\end{subfigure}
	\begin{subfigure}[b]{0.49\linewidth}
		\centering
		\includegraphics[width=\linewidth,trim=15 2 25 25,clip]{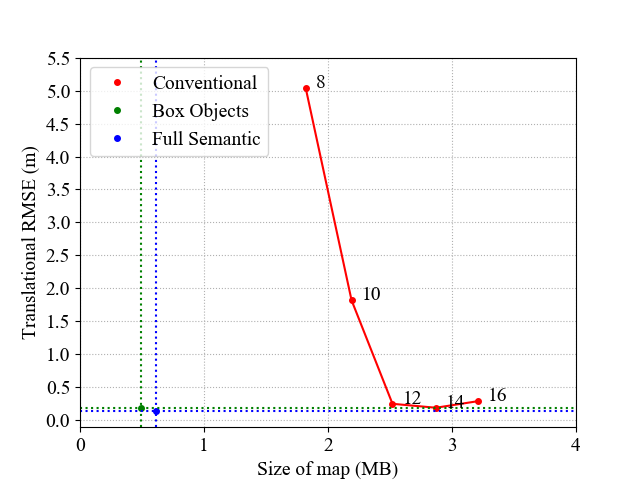}
		\caption{Hangzhou-2}
		\label{fig:10_result}
	\end{subfigure}
	\caption{Self-relocalization results on four KAIST and two Hangzhou sequences. Red: The results with sparsified conventional maps
 	containing point descriptors
    under different set cover parameters $K$. Blue: Proposed method without lanes. Green: Proposed method.}
	\label{fig:kaist_results}
\end{figure}

\textbf{Self-Relocalization.}
In this test, we run each sequence twice for mapping and map-aided localization, respectively. Fig.~\ref{fig:kaist_results} lists the results on tested sequences.
The semantic map generated by our proposed method requires much less file size while providing a competitive localization performance. For example, the semantic map of KAIST-38 (Fig.~\ref{fig:38_result}) spends 1.54 MB disk space to reach 0.46 meters localization precision, while the conventional map requires 7.60 MB for 0.75 meters by setting $K = 8$.
Adjusting the sparsification parameter $K$ cannot reach the same size-efficiency of our method.
Tests on Hangzhou sequences also reflect the same trend. Overall, the semantic maps in our proposed method are nearly 5 times smaller than those set cover point feature maps to reach similar localization performance.

\begin{table}[ht]
	\centering
	\caption{Cross-validation results. We report the T.RMSE (T) in meters, map size in MB (Size), preserving percentage ($D$), set cover parameter ($K$), and the map size ratio (Sz.Ratio) between two compared methods and ours.}
	\setlength{\tabcolsep}{2pt}
	\begin{tabular}{ccc|ccc|cc|cc}
		\hline
		\multicolumn{3}{c|}{Maplab~\cite{schneider2018maplab}} & \multicolumn{3}{c|}{Conventional} & \multicolumn{2}{c|}{Box objects only} & \multicolumn{2}{c}{Full Semantic} \\ \hline
		T      & Size  & $D$  & T     & Size  & $K$  & \makecell[c]{T \\ Size} & Sz.Ratio & \makecell[c]{T \\ Size} & Sz.Ratio    \\ \hline
		\multicolumn{10}{c}{KAIST-38 localization on KAIST-28 Map} \\ \hline
		\multicolumn{3}{c|}{\multirow{2}{*}{Failed}} & \multicolumn{3}{c|}{\multirow{2}{*}{Failed}} & \multirow{2}{*}{\makecell[c]{1.49 \\ 1.31}} & - & \multirow{2}{*}{\makecell[c]{\textbf{1.40} \\ 1.83}} & - \\
		& & & & & & & & & \\ \hline
        \multicolumn{10}{c}{KAIST-28 localization on KAIST-38 Map} \\ \hline
		12.85  & 35.77 & 20\% & 7.88  & 43.60 & 60   & \multirow{4}{*}{\makecell[c]{\textbf{0.96} \\ \\ 1.54}}  & $\times 23,28$ & \multirow{4}{*}{\makecell[c]{0.99 \\ \\ 2.12}} & $\times 17,21$ \\
		   4.28    &   54.12    & 35\% & 10.97 & 57.29 & 80   & & $\times 35,37$ & & $\times 26,27$ \\
		2.86   & 65.92 & 50\% & 0.93  & 69.50 & 100  & & $\times 43,45$ & & $\times 31,33$ \\
		2.49   & 94.88 & full & 0.86  & 94.88 & full & & $\times 62,62$ & & $\times 45,45$ \\ \hline
		\multicolumn{10}{c}{Hangzhou-2 localization on Hangzhou-1 Map} \\ \hline
		8.89   & 11.27 & 20\% & 8.55  & 8.04  & 30   & \multirow{4}{*}{\makecell[c]{0.29 \\ \\ 0.56}}  & $\times 20,14$ & \multirow{4}{*}{\makecell[c]{\textbf{0.18} \\ \\ 0.62}} & $\times 18,13$\\
		  3.54     &  16.86     & 35\% & 0.89  & 10.37 & 40   & & $\times 30,18$ & & $\times 27,17$ \\
		1.59   & 19.66 & 50\% & 0.38  & 12.40 & 50   & & $\times 35,22$ & & $\times 32,20$ \\
		0.74   & 29.20 & full & 2.54  & 14.00 & 60   & & $\times 52,25$ & & $\times 47,23$ \\ \hline
	\end{tabular}
	\label{table:cross_all}
\end{table}

\textbf{Cross-Relocalization.}
We conduct the cross-relocalization experiments with both conventional maps and a state-of-the-art system Maplab~\cite{schneider2018maplab}. Same with our two-stage manner, Maplab is also an open source mapping and localization framework. Since its original version does not utilize GNSS measurements, we transform those maps constructed by our baseline method~\cite{chen2019perception,zhang2019localization} to enhance the quality of localization summary maps. Furthermore, we test its included map sparsification method~\cite{dymczyk2015keep} for compression by different percentages $D$ of point landmarks preservation.

Table~\ref{table:cross_all} shows such reciprocal cross-relocalization results. As can be seen, our proposed method with semantic maps containing box objects and road lanes accomplishes a successful localization on all cross-validation tests. For conventional maps, we test with multiple settings including a full-size map. Unfortunately, the localization consistently fails on running the KAIST-38 sequence on the map of 28 sequence, due to the severe change of visual features captured in different seasons and time periods.
Hence, the set cover sparsification needs to apply a larger $K$ value for cross-relocalization to maintain sufficient reliable features. For Maplab, The resultant map size is larger than expected, since the preserved informative 3D points are usually associated with more local visual descriptors. In contrast, our proposed method relies on deep features extracted from standardized and persistent road elements.
In conclusion, such semantic replacement outperforms both conventional maps and Maplab by a larger margin on cross-localization, and experiments on Hangzhou sequences reflect the same trend, which shows the superiority in both compactness and localization accuracy.

\textbf{Modular and Efficiency Analysis.} Table.~\ref{table:state_stat} presents the statistics of two {\em tracking} states involved in our localization. In most cases, our method runs in the {\em tight tracking} mode, which reflects a better localization accuracy. Table~\ref{table:point_stat} shows the average number of points used from different semantic entities. As dash cams can observe and detect valid pole and sign objects relatively far than ground objects, these objects contribute the most to the localization.

We also analyze the time consumption for semantic object association in different localization modes. During online localization, both the perception and the map query module runs in a standalone thread. In the perception module, it takes 17.6 ms / 693 MB for lanes, 7.5 ms / 422 MB for corners, and 17.3ms / 1177 MB for other objects and keypoints if individually detected respectively. In the map query module, the average time consumption of the \emph{local search} mode is around 300 ms, while for the dominant \emph{loose tracking} or \emph{tight tracking} modes it reduces to less than 1 ms. The overall time consumption is lower than our designed localization query frequency (1Hz).

\begin{table}[ht]
	\centering
	\caption{The number of attempts, percentage, and T.RMSE in meters of different tracking states.}
	\begin{tabular}{c|rrc|rrc}
		\hline
		         & \multicolumn{3}{c|}{Loose Tracking} & \multicolumn{3}{c}{Tight Tracking} \\
		         & Num. &  \%   & T.RMSE        & Num.   & \%    & T.RMSE              \\ \hline
		         \multicolumn{7}{c}{KAIST Dataset} \\ \hline
		38 on 28 & 157  & 9.3   & 1.01          & 1533   & 90.7 & \textbf{0.81}     \\
		28 on 38 & 264  & 17.2  & 1.04          & 1272   & 82.8 & \textbf{0.75}     \\
		26       & 20   & 3.9   & \textbf{0.15} & 488    & 96.1 & 0.17              \\
		28       & 161  & 10.8  & 1.13          & 1330   & 89.2 & \textbf{0.47}     \\
		38       & 154  & 7.9   & 0.89          & 1796   & 92.1 & \textbf{0.37}     \\
		39       & 111  & 7.4   & 0.50          & 1386   & 92.6 & \textbf{0.13}     \\ \hline
		\multicolumn{7}{c}{Hangzhou Dataset} \\ \hline
		2 on 1  & 42    & 10.6  & 0.30          & 354    & 89.4 & \textbf{0.16}      \\
		1       & 80    & 14.1  & 0.39          & 487    & 85.9 & \textbf{0.27}      \\
		2       & 35    & 8.5   & 0.27          & 377    & 91.5 & \textbf{0.17}      \\ \hline
	\end{tabular}
	\label{table:state_stat}
\end{table}

\begin{table}[ht]
	\centering
	\small
	\caption{The summarization about the average number of points used from semantic entities under different experimental settings. We count the descriptor-less classical feature points (CP), deep points from poles and signs (P+S), ground (GND) boxed objects, and lane entities.}
	\label{table:state_28_on_38}
	\begin{tabular}{c|c|ccc|cc}
		\hline
		&       & \multicolumn{3}{c|}{Box Objects} & \multicolumn{2}{c}{Lane} \\
		&       & CP  & P+S  & GND  & Corners       & Samples      \\ \hline
		\multirow{2}{*}{KAIST}    & self  & 3.10    & 4.67    & 1.99    & 1.65         & 178.61      \\
		& cross & 0.85    & 3.87    & 1.65    & 3.11         & 171.67      \\ \hline
		\multirow{2}{*}{Hangzhou} & self  & 0.82    & 6.83    & 3.52    & 3.45            & 64.35       \\
		& cross & 0.70    & 8.12    & 3.51    & 2.96           & 68.14       \\ \hline
	\end{tabular}
	\label{table:point_stat}
\end{table}

\textbf{Stability of Semantic Mapping}
As experiments above mainly demonstrate the effectiveness during online localization, we additionally perform a quantitative evaluation on the mapping quality, by comparing trajectories during both conventional mapping and semantic mapping to their corresponding dataset baseline, i.e., the GNSS trajectory of Hangzhou sequences and provided ground truth trajectory of KAIST, for clarifying the stability of involving such semantic features. Table~\ref{table:mapping_quality} shows the mapping errors in both classical and semantic mapping methods. This suggests that introducing semantic objects does not obviously influence the mapping quality.

\begin{table}[ht]
	\centering
	\caption{Mapping T.RMSE in meters compared to the given ground truth trajectory.}
	\begin{tabular}{c|cccc|cc}
	    \hline
	    		  & \multicolumn{4}{c|}{KAIST}     & \multicolumn{2}{c}{Hangzhou}  \\
	              & 26       & 28       & 38       & 39       & 1     & 2          \\ \hline
	    Classical & 1.75     & 2.69     & 2.61     & 2.41     & 0.81  & 0.85       \\
	    Semantic  & 1.77     & 2.67     & 2.63     & 2.43     & 0.82  & 0.84       \\
	    \hline
	\end{tabular}
	\label{table:mapping_quality}
\end{table}

\section{Conclusion}
\label{sec:conclude}

In this paper, we presented a semantic mapping and localization pipeline. Entities including poles, signs, and road lanes, are detected and parameterized to form a compact  semantic map for efficient and accurate localization.




\bibliographystyle{IEEEtran}
\bibliography{reference}



\end{document}